\documentclass[conference]{IEEEtran}
\IEEEoverridecommandlockouts
\usepackage{cite}
\usepackage{amsmath}
\usepackage{amssymb}
\usepackage{amsthm}
\usepackage{ amsfonts}
\usepackage{algorithmic}
\usepackage{graphicx}
\usepackage{textcomp}
\usepackage{url}
\usepackage{color}

\def\BibTeX{{\rm B\kern-.05em{\sc i\kern-.025em b}\kern-.08em
    T\kern-.1667em\lower.7ex\hbox{E}\kern-.125emX}}
\begin{document}

\title{A VEST of the Pseudoinverse Learning Algorithm \\
{\footnotesize \textsuperscript{*}}
\thanks{This work is fully supported by the grants from the Joint Re-search Fund
in Astronomy (Grant No. U1531242) under cooperative agreement between the National Natural
Science Foundation of China (NSFC) and Chinese Academy of Sciences (CAS).}
}

\author{\IEEEauthorblockN{Ping Guo}
\IEEEauthorblockA{\textit{School of Systems Science} \\
\textit{Beijing Normal University}\\
Beijing 100875,  China }
}
%\and

%\and
%\and
%\and

\maketitle

\begin{abstract}
In this paper, we briefly review the basic scheme of the pseudoinverse learning (PIL) algorithm and present some discussions on the PIL, as well as its variants. The PIL algorithm, first presented in 1995, is a non-gradient descent and non-iterative learning algorithm for multi-layer neural networks and has several advantages compared with gradient descent based algorithms. Some new viewpoints to PIL algorithm are  presented, and several common pitfalls in practical implementation
of the neural network learning task are
also addressed.  In addition, we  show that so called extreme learning machine  is a \textcolor{blue}{ {\bf \underline v}}ariant cr\textcolor{blue}{\bf \underline e}ated by \textcolor{blue}{\bf \underline s}imple name al\textcolor{blue}{\bf \underline t}ernation (VEST) of the PIL algorithm for single hidden layer feedforward neural networks. 
\end{abstract}
\vspace{0.2cm}


\section{Introduction}
\label{intro}
Multi-layer perceptron (MLP) is a kind of feedforward neural networks, which is most studied in mid-eighties of the last century. With more than three hidden layers, MLP is called deep neural network (DNN), and training DNN is a deep learning procedure. Now MLP has already been found to be successful for various supervised learning tasks. Both theoretical and empirical studies have shown that the MLP is of powerful capabilities for pattern classification and universal approximation \cite{Bishop1995}.  When there are few hidden layers, weight parameters of the network can be learned by the gradient descent learning algorithm, namely the well-known error back propagation (BP) algorithm \cite{Rumelhart1986Learning}\cite{Lecun1998Gradient}. As we have known, the BP algorithm has several disadvantages. It usually has a poor convergence rate and sometimes falls into local minima \cite{Barnard1992}. The selection of hyperparameters in the BP algorithm, such as learning rate and momentum constant, is often crucial for the success of the algorithm.

In order to solve these problems arisen in the BP algorithm, Guo {\it et al}  \cite{Guo1995} proposed a non-gradient descent algorithm, and later named this algorithm as the pseudoinverse learning (PIL) algorithm \cite{Guo2001}. Unlike the BP algorithm, the PIL algorithm could exactly calculate the network weights, rather than to find weights with iterative optimization. The PIL algorithm adopts only generalized linear algebraic methods, e.g., pseudoinverse operations and matrix inner products. Moreover, PIL does not need to explicitly set any control parameters, which were usually specified by users empirically. 

\section{Basic Scheme of the Pseudoinverse Learning Algorithm}
\label{sec:1}

\begin{figure}[htbp]
\centerline{\includegraphics[width=8cm, height=7cm]{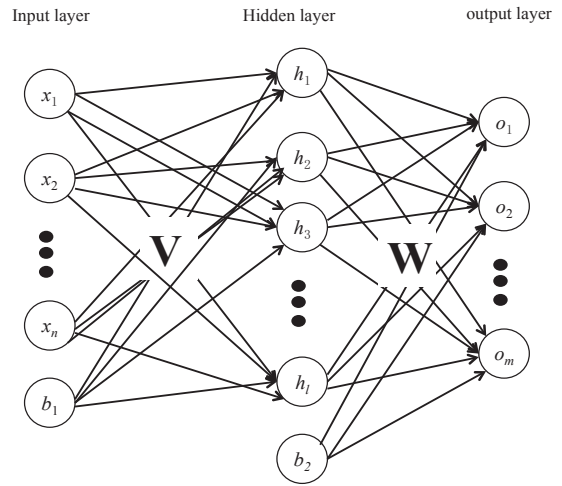}}
\caption{A schematic diagram of single hidden layer neural networks.}
\label{fig1}
\end{figure}

BP algorithm is a kind of gradient decent algorithm. Evidently, it is a great discovery in neural network learning, but it has a few of disadvantages, such as slow convergence or local minima, which most researchers knew about. Non-gradient descent based algorithms have been considered as alternative approaches, especially in the late 90 or early 2000, among which the PIL algorithm is a success one \cite{Guo1995}\cite{Guo2001}, especially for multi-layer perceptron (or multi-layer neural networks). Following we will take a single hidden layer neural network (SHLN) as a example to review the learning algorithms. 

Figure (\ref{fig1}) is a schematic diagram of single hidden layer neural networks. 

The SHLN shown in Fig. (\ref{fig1}) has three layers, including  one input layer, one output layer and one hidden layer.  There are $n+1$ neurons in input layer, $l+1$ neurons in hidden layer, and $m$ neurons in output layer.  We use ${\bf x}=(x_1,  x_2, ..., x_n)$ to express $n$ dimensional input vector,   and ${\bf o}=(o_1,  o_2, ..., o_m)$ stands for $m$ dimensional output vector.  $b_1$ in input layer is called bias neuron,  and $b_2$ is hidden layer bias neuron. ${\bf V}$ is the weight matrix connecting input and hidden neurons, and ${\bf W}$ is the weight matrix connecting the hidden and output neurons. The network mapping function is expressed as,

\begin{eqnarray}
{o_k}&=& f_k({\bf x, \Theta}) \nonumber \\
          &=& \sigma(\sum _{j=1}^{l}W_{k, j}g_{j} +b_2), \nonumber \\
 g_j &=& \sigma(\sum _{i=1}^{n}V_{j, i}x_i +b_1).
\end{eqnarray}

Where ${\bf \Theta}$ stands for the network parameter group, including connecting weights ${\bf W, V}$, and bias neurons $b_1, b_2$. 
While $\sigma(\cdot)$ is an activation function, most used function type including sigmoid, hyperbolic, step, radial basis function, and so on.

From above equation, we can see that ${\bf{g} }({\bf Vx +b}_1)$ is the hidden layer output, and ${\bf{f} }({\bf Wg +b}_2)$ is the last layer output.
If we let
\begin{eqnarray*}
\sum _{i=1}^{n}V_{j, i}x_i +b_1 =  \sum _{i=0}^{n}V_{j, i}x_i  \\
\mathtt {with}\,\,\,  V_{j, 0}=b_1,  x_0=1, 
\end{eqnarray*}

and
\begin{eqnarray*}
\sum _{j=1}^{l}W_{k, j}g_{j} +b_2=\sum _{j=0}^{l}W_{k, j}g_{j}\\
\mathtt{with} \,\,\,  W_{k, 0}=b_2,  g_0=1,
\end{eqnarray*} 
now matrix 
${\bf V, W} $   become augmented matrix.  In the literatures,  bias neuron is used to prevent zero input vector to destroy  weight updating in sequentially training, the value of the bias usual is set to be $+1$, but some researchers also took it as a variable.  For mathematical expression concise,  hidden layer bias neuron $b_2$ often is omitted.  Please note that in Ref. \cite{Huang2006}  $b_1$ is called the threshold of the $i$th hidden node, but it has no role for threshold operation.

When data set $D=\{{\bf x}^i, {\bf t}^i \}$ is given, training the network is to find the weigh parameters ${\bf V, W}$ with minimizing cost function. 
The cost function, also known as the loss function, or error function, is used to measure the difference between the actual outputs and the expected outputs of the network. There are many forms of the error function, which is determined by the probability distribution of the error. When the error distribution is the Gaussian distribution, the cost function is the sum of the squared error (SSE). 

\begin{equation}
E=\frac{1}{2N}\sum _{i=1}^{N}\sum _{j=1}^{m}\| f_j({\bf x}^i, {\bf \Theta}) - t_j^i \| ^2.
\label{sse2}
\end{equation}

For simplifying, we can write this system error function in matrix form,

\begin{equation}
E=\frac{1}{2N}\| {\bf O -T }\|_F ^2.
\label{e2}
\end{equation}

Where subscript $F$ stands for  Frobenius norm.

The purpose of neural network training is to find the weight parameters to minimize the cost function. Traditional learning algorithm is BP algorithm. 
As mentioned earlier, learning algorithm related hyperparameters, such as learning rate and momentum constant,  need to be selected by users in BP algorithms.  The choices of hyperparameters are more difficult for most beginners of neural network research.

In order to overcome the drawbacks of the BP algorithm,  Guo {\em et al} \cite{Guo1995}  proposed the PIL algorithm to train a SHLN in 1995. In that work, the activation function is taken as the hyperbolic function $Tanh(á)$.  Minimizing following error function to find weight parameter matrix,
\begin{equation}
minimize \| {\bf YW-B }\|^2,
\label{minerr}
\end{equation}
where ${\bf Y} = Tanh ( {\bf XV}) $  is the output matrix of the hidden layer,  ${\bf X}$ is the input matrix consisting of $N$ input vectors as its rows and $d=n+1$ columns as input vector dimension $n$ plus 1,  ${\bf B} = ArcTanh ({\bf T})$,  and ${\bf T}$ is the target label matrix which consists of $N$ label vectors as its rows and $m$ columns as target vector dimension. 

Eq. (\ref{minerr}) is formally a problem of least squares in linear algebra. However, only the ${\bf B}$ matrix is known at present. Weight matrices {\bf V} and {\bf W} are not yet known. The task of the network learning is to find these matrices.  According the theorems in linear algebra, a formal solution to {\bf W} in Eq. (\ref{minerr}) is
$ {\bf W = Y}^+ {\bf B}$ , where ${\bf Y}^+$ is the pseudoinverse of  ${\bf Y}$. To bring this formal solution into Eq. (\ref{minerr}), we can also get the following mathematical form:

\begin{equation}
\bf YW-B=YY^+ B-B=0.	
\label{eq02}
\end{equation}

If Eq. (\ref {eq02}) holds,  an intuitive explanation is that ${\bf YY^+=I }$ will satisfy the requirement. So if ${\bf Y}^+ $ is the final solution, ${\bf Y}$  should be a full rank matrix. With this new objective function, we can adopt some methods to set input matrix ${\bf V}$ so as to make hidden layer output matrix ${\bf Y}$  to approach the full rank.  A simple way to set ${\bf V}$  to be a random value matrix could reach this goal with nonlinear transformation, or we can set ${\bf V}$  as the pseudoinverse of the input matrix ${\bf X}$.      

As we known, in SHLN, the number of hidden neurons $l$ is a hyperparameter of  the neural network architecture. When a given problem is formulated, the number of neurons in the input layer is determined by the dimensionality of the input data, and the number of neurons in the output layer depends on the specific problem. The number of hidden layer neurons is only one hyperparameter for SHLN and perhaps it the hardest problem for most beginners (see \cite{NNFAQ} for more details).  In the work of Guo {\it et al}  \cite{Guo1995},  the $l$ is set be $N$ for the purpose of exact learning. 

Hence, we have a fast learning algorithm which computes the weight matrix instead of iterative approach. For
most problem, it only needs one step to reach the perfect learning. The algorithm can be  summarized as follows:

\vspace{0.3cm} 
{\bf Algorithm PIL:}   Given a date set, we draw $N$ pair samples  $D=\{{\bf x}^i, {\bf t}^i \}_{i=1}^{N}$ as  the training set,  activation function $Tanh(\cdot)$, and set hidden neuron number be $N$,
\begin{itemize}
\item[] {\it Step 1}: Compute $V = Pseudoinverse[X]$  and hidden layer output matrix $Y=Tanh(XV)$,
\item[] {\it Step 2}: Compute $Y^+=Pseudoinverse[Y]$ and 

$B = ArcTanh[T]$, 
\item[] {\it Step 3}: Compute   the output weight matrix 

$W=Y^+B$.
\end{itemize}

And the network output ${\bf o}= Tanh(Tanh({\bf xV)W})$.

As we known, the common practice is random initialization of weight parameters in BP algorithm, then with delta learning rule to update weight matrix. While in PIL algorithm,
weight parameters are computed with pseudoinverse solution, and do not need to be adjusted further.  In the work of Guo {\it et al}  \cite{Guo1995}, randomly set input weight  
$V$ also has been investigated. Following sentence is copied from Ref. \cite{Guo1995}:

\begin{quotation}
 {\em A simple method is to set V as a random n by N matrix.  In practice, this is not a  proper method.  As we
mentioned above, we use $Tanh[\cdot]$ as the activate function. If the matrix Z =XV contains elements of
large values,  it will result in complex numbers which was not desirable.  So it is better to choose a proper
matrix V so that Z has no elements of large value. One way is to set the values of the elements as small
as possible. }
\end{quotation}

When writing  above description in following form, we can regard it as a variant of  the PIL algorithm:

\vspace{0.3cm} 
{\bf Algorithm PIL0:}   Given a date set, we draw $N$ pair samples  $D=\{{\bf x}^i, {\bf t}^i \}_{i=1}^{N}$ as  the training set,  activation function $Tanh(\cdot)$, and set hidden neuron number  $l=N$,

\begin{itemize}
\item[] {\it Step 1}: Randomly assign input weight matrix V (the  element values in V should in a small interval, \\
 say [-1, +1]),
 \item[] {\it Step 2}: Compute the hidden layer output matrix \\
 $Y=Tanh(XV)$, 
\item[] {\it Step 3}:  Compute  the output weight matrix $W$

          $W=Y^+B$.

           where B=ArcTanh(T)
\end{itemize}

And the network output ${\bf o}= Tanh(Tanh({\bf xV)W})$.

\vspace{0.3cm}

{\bf Remarks}

\begin{enumerate}
\item Set hidden neuron number $l=N$ is for exact learning,  if the training error is allowed, we can set hidden neuron number $l<N$.
\item Activation function can be taken any nonlinear transformation function, such as sigmoid, Gaussian kernel, and so on.
\item When last layer activation function is taken as a linear function, $W=Y^+T$ \cite{Guo2001}.
\end{enumerate}

Recently, we find that Huang {\it et al} \cite{Huang2004}\cite{Huang2006}  created a name called extreme learning machine (ELM),  compared ELM algorithm described in \cite{Huang2004}\cite{Huang2006},  we can easily find that it  
  is exactly the same with our PIL0 algorithm for SHLN in learning scheme.  So we think that ELM algorithm  is a {\bf \underline v}ariant cr{\bf \underline e}ated by {\bf \underline s}imple name 
  al{\bf \underline t}ernation (VEST) of the PIL algorithm.

\section{ PIL Variants}

For a number of data set,   the PIL algorithm for SHLN can reach accurate learning. But for some other data, the learning accuracy of  SHLN cannot meet the high precision requirements 
if we only simply assign input weight matrix V as the pseudoinverse of X, or a random value matrix without any constraints.  In 2001, Guo {\it et al} \cite{Guo2001}  proposed a new solution that extended the neural network architecture from a single hidden layer to multiple hidden layers.  Later in 2003, Guo {\it et al} extended their work and published in the journal of {\em Neurocomputing} (copyright 2003) \cite{Guo2004}.

The PIL algorithm for multilayer neural network is summarized as follows:

\vspace{0.3cm} 
{\bf Algorithm ePIL:}   Given a date set, we draw $N$ pair samples  $D=\{{\bf x}^i, {\bf t}^i \}_{i=1}^{N}$ as  the training set,  activation function $\sigma(\cdot)$, and set hidden neuron number be $N$,
\begin{itemize}
\item[] {\it Step 1}: Compute $(Y^0)^+ = Pseudoinverse[X]$,
\item[] {\it Step 2}: Compute $\|Y^l (Y^l)^+ - I\|^2 $. If it is less than the given error $E$, go to step 5. If not, go on to the next step.
\item[] {\it Step 3}:  Let $W^l=(Y^l)^+$.  Feed forward the result to next layer, and compute  $ Y^{l+1} =\sigma (Y^lW^l) $.
\item[] {\it Step 4}:  Compute   $ (Y^{l+1})^+ =Pseudoinverse(Y^{l+1})$, set $l \leftarrow l+1$, and go to step 2.
\item[] {\it Step 5}:  Let final layer  output weight matrix \\
$W^L=(Y^L)^+ T$. 
\end{itemize}

And the network output \\
\begin{equation}
 {\bf o}= \sigma(...\sigma(\sigma({\bf xW^0)W^1)...)W}^{L}.
\label{dnn}
\end{equation}

\vspace{0.3cm} 

There  are some new viewpoints to PIL is as follow:

{\bf Remarks}
\begin{enumerate}
\item Eq.(\ref{dnn}) showed a deep neural network architecture.
\item The depth of this DNN is dynamical growth and is data dependent.
\item If  stoped at $\|Y^l (Y^l)^+ - I\|^2 =0$,  we get an identity orthogonal projector  $P=Y^L (Y^L)^+ $
\item When we let $T=I$,  the PIL algorithm is an unsupervised learning algorithm, it realizes vector normalization in high dimensional space.
\end{enumerate}

In the work of Guo {\it et al}  \cite{Guo2004}, another variant of PIL algorithm is discussed. It stated in the discussion section of   Ref. \cite{Guo2004}:

\begin{quotation}
 {\em  ``But if we intend to reduce the network complexity, we can add a same-dimension Gaussian noise matrix to perturb the transformed matrix in step 4 of the PIL algorithm.
  The inverse function of the perturbed matrix will exist with probability one because the noise is an identical and independent distribution. In such a strategy, we can constrain the hidden layers to at most two to reach the perfect learning''}.
 
\end{quotation}
 
Writing above description with mathematical algorithm form:

\vspace{0.3cm} 
{\bf Algorithm PIL1:}   Given a date set, we draw $N$ pair samples  $D=\{{\bf x}^i, {\bf t}^i \}_{i=1}^{N}$ as  the training set,  activation function $\sigma(\cdot)$,  set hidden neuron number be $N$, and set  Gaussian noise perturbation   matrix be $\tilde{G}$.
\begin{itemize}
\item[] {\it Step 1}: Compute $(Y^0)^+ = Pseudoinverse[X]$,
\item[] {\it Step 2}: Compute $\|Y^l (Y^l)^+ - I\|^2 $. If it is less than the given error $E$, go to step 5. If not, go on to the next step.
\item[] {\it Step 3}:  Let $W^l=(Y^l)^+  +\tilde{G}$.  Feed forward the result to next layer, and compute  $ Y^{l+1} =\sigma (Y^lW^l) $.
\item[] {\it Step 4}:  Compute   $ (Y^{l+1})^+ =Pseudoinverse(Y^{l+1})$, set $l \leftarrow l+1$, and go to step 2.
\item[] {\it Step 5}:  Let final layer  output weight matrix \\
$W^L=(Y^L)^+ T$. 
\end{itemize}

{\bf Remark}

\begin{enumerate}
\item[] Weight matrix adding noise is equivalent  to input matrix adding noise, while training with noise is a kind of regularization \cite{BishopNC1995}.
\end{enumerate}

\subsection{VEST Analysis}

\subsubsection{Neural Network Architecture}
\label{leN}
For a MLP,  the number of hidden layers and the number of hidden layer neurons belong to the architecture hyperparameters, there is no theory to guide how to select these hyperparameters.  For a SHLN, only one architecture hyperparameter is the number of hidden layer neurons,  and perhaps it is also a difficult problem for most beginners. 
In Ref.\cite{Guo1995},  Guo {\it et al} detailed the reason for choosing the number of hidden neurons as follows:

``{\em  From the above definitions (B is an $N\times m$ matrix, Y is an $N\times l$ matrix), we know $Y^+ $ is an $l \times  N$ matrix.  Based on linear algebra we know that if $l<N$, it is impossible that Y has a right inverse [6]\footnote{[6] in Ref.\cite{Guo1995}  is ``Ben Noble and James W. Daniel, {\it Applied Linear Algebra}, Prentice-Hall, Englewood Cliffs, NJ, 1988.} ''. This requires that at least $l=N$ if we hope to get inverse. (of course, we can choose $l>N$, but this will increase computation time)}''.  It is explicit that a suggestion on the selection of the number of hidden neurons is given in the PIL.  In contrast, a specific value of the number of hidden neurons was not given in the ELM papers, and it did not give any effective suggestion for this hyperparameter selection beyond saying 
``random nodes''.  In fact, in Ref.\cite{Huang2004}\cite{Huang2006},  they just follow those discussions about the number of hidden neurons should be set to $N$,  no other specific value was suggested.

As for activation function,  the most commonly used  functions in MLP are  sigmoidal function and hyperbolic tangent function. 
Here are some discussions on activation functions: In \cite{Guo2004}, it stated, {\em From the learning procedure, it is obvious that no differentiable activate function is needed. We only require that the activate function can perform nonlinear transform to raise the rank of the weight matrix.}  In \cite{Huang2004}, it  repeated our no differentiable activate function statement and gave no explanations, ``{\em Unlike the traditional classic gradient-based learning algorithms which only work for differentiable activation functions, the ELM learning algorithm can be used to train SLFNs with non-differentiable activation functions}''.  From this point, it is  doubt that authors of the paper \cite{Huang2004} have read our papers previously.  While in \cite{Huang2006}, similar statement appeared in the discussion section, ``{\em Unlike the traditional classic gradient-based learning algorithms which only work for differentiable activation functions, as easily observed the ELM learning algorithm could be used to train SLFNs with many nondifferentiable activation functions}''.

     However, still in \cite{Huang2006},  in order to prove theorem 2.1, the authors required the activation function is infinitely differentiable. It also stated that Ò{\em This paper rigorously proves that for any infinitely differentiable activation function SLFNs with N hidden nodes can learn N distinct samples exactly and SLFNs may require less than N hidden nodes if learning error is allowed}.Ó  The question is that why activation function from nondifferentiable in 2004\cite{Huang2004} becomes infinitely differentiable in 2006\cite{Huang2006},  in \cite{Huang2006}, authors did not give any explanations.

\subsubsection{Weight Parameters}

Weight parameters are a set of important parameters in neural network. The initialization of weight parameters directly determines generalization performance and convergence rate of neural networks when the BP algorithm is used in network learning. After a neural network architecture is designed, it is common to adopt the BP algorithm to find these weight parameters. When applying the BP algorithm, the traditional way to initialize the weight parameters is to randomly set these values.  In \cite{Guo1995}\cite{Guo2001}\cite{Guo2004}, the weight matrix W, which connect hidden and  output neurons, is computed with pseudoinverse $W=Y^+T$. While  in \cite{Huang2004}\cite{Huang2006},  the same method is described with  that in PIL.  As for weight matrix, V or $W^0$, which connects input and hidden neurons, three methods have been investigated by Guo {\em et al}: 
\begin{enumerate}
\item[(1)] It can be set as pseudoinverse of input matrix as presented in \cite{Guo1995}\cite{Guo2001}\cite{Guo2004}; 
\item[(2)] It can be initialized randomly without further tuning, as stated  in  \cite{Guo1995};  
\item[(3)] It can be set as pseudoinverse of input matrix with additive Gaussian noise, as stated  in \cite{Guo2004}.
\end{enumerate}
From these facts, it is clear that random weight generation in \cite{Huang2004}\cite{Huang2006} is just one of the choices within PIL.

\subsection{Other Variants}

Wang and Wan pointed out in their Comments \cite{Wang2008}:  ``{\em The output weights can be adjusted in one of the following ways: 1) using pseudoinverse (also known as Moore--Penrose generalized inverse); 2) incrementally (at each iteration, a new random hidden neuron is added); or 3) online sequentially (as new data arrive in real-time applications)}''.  We note the following: 

1) is the exact same as that in PIL papers.  

2) is a simple extension of Griville's theorem (add neuron) and bordering algorithm (delete neuron) which is discussed in \cite{Guo2001}\cite{Guo2004}.  

3) is also a simple extension of Griville's theorem which is discussed in \cite{Guo2001}\cite{Guo2004}. 

Wang and Wan also pointed out, ``{\em In conclusion, feedforward networks (both RBF and MLP) with randomly fixed hidden neurons (RHN) have previously been proposed and discussed by other authors in papers and textbooks. These RHN networks have been shown, both theoretically and experimentally, to be fast and accurate. Hence, it is not necessary to introduce a new name `ELM'. } ''  Here we notice that  in \cite{Huang2004}\cite{Huang2006}, only the MLP feedforward network with sigmoidal activation function is referred, and they only discussed that with $N$ hidden neurons the training error can reach zero, as that discussed in PIL papers.  Hence our discussions at here are restricted to only these two papers, without referring to Huang{'}s other papers\footnote{More discussions can be found at https://elmorigin.weebly.com},  or other PIL's variants after year 2004. Under this restriction, we can see that from the learning scheme to concrete methods, the authors of papers \cite{Huang2004}\cite{Huang2006} followed our PIL work, created a new name called ELM, with nothing new except leaving the hardest work on the selection of the number of hidden nodes to users. From these discussion, it is easy to know that the ELM is a VEST of the PIL algorithm.

\section{Some Statements}

In this section,  we will  point out some incorrect statements and false claims in \cite{Huang2004}\cite{Huang2006}.

\subsection{Data Interval}
The theorem 2.1 in \cite{Huang2006} claimed that for the activation function $g$ which is infinitely differentiable in any interval, with $N$ training samples, randomly chosen input weights and hidden layer bias from any intervals, the hidden layer output matrix H is invertible. However, this theorem is incorrect. It is a common pitfall that randomly draws data in any interval, 
for example, if we take Tanh(x) as the activation function,  when we randomly chose the input weights and bias from any intervals, the hidden output matrix H is often not invertible as it is investigated in \cite{Guo1995}.  The reason is that those very big or very small values in input weight matrix will make Tanh(x) function saturation, resulting in these elements assuming the same values in matrix H (rank defective) and H being non-invertible in numerical simulations.  

In fact,  strictly speaking, we have,

\vspace{0.2cm}
{\em Theorem 1}.  For any bounded activation function $g$,  if no constraint to its input intervals, the hidden layer output matrix H is not always invertible.

\begin{proof}
 Given a standard SHLN,  define interval $(a/\epsilon, b/\epsilon)$,  with $\lim \epsilon \rightarrow 0$. When we randomly chosen $w_i,  b_i$  data from $b>a>0$ interval, or $b<a<0$ interval,  the elements in hidden layer output matrix H will be its boundary value when $\lim \epsilon \rightarrow 0$. For example, elements in H=sigmoid(WX) will be all 1 or 0 for finite value training data set X.  The rank of the H matrix is 1 and it is not full when $N>1$,  then it is NOT invertible with probability one. 
\end{proof}

In practical implement an algorithm in computing,  it is a pitfall if no constraint is considered. The IEEE floating-point standard  specifies the positive and negative infinity values.  The single precision effective floating point range is about $ \approx \pm  10^{38.53}$,  the  numbers great than positive or less than negative $10^{38.53}$  will be regards as infinity.

Another example in \cite{Wang2017} is when we choose randomly the input weights and biases in [-1,1], it will be found that the rank of hidden output matrix H is NOT full:

 Let $f(x)$ be defined over [0,1],
\begin{eqnarray*}
f(x) =0.2\exp\{-(10x-4)^2\}\\
 +0.5\exp\{-(80x-40)^2\}   \\
               +0.3\exp\{-(80x-20)^2\}.
\end{eqnarray*}

If this function is approximated with SHLN, it will be found that the theorem 2.1 in \cite{Huang2006} is incorrect.

\subsection{Training Error}

The theorem 2.2 in \cite{Huang2006} claimed that if any small positive value $\epsilon >0$ is given, there exists $\tilde{N}\leq N$, for any random input weight values, 
$\| H_{N\times \tilde{N}}\beta _{\tilde{N}\times m}-T_{N\times m}\| <\epsilon $.   And the proof was given only the case of  $ tilde{N}=N$ according to Theorem 2.1.
Here we do not consider whether theorem 2.1 is correct or not, but discuss only the case of  $ \tilde{N}=N$ first.  From the linear algebra textbook it is known that if H is a square full rank matrix, the inverse of H exists and exact learning can be reached as we discussed in \cite{Guo1995}. So for the case of $\tilde{N}=N$,  it does
not need to prove training error $\epsilon $  can be any small
value again because     $\epsilon $  can be zero.  Even if it needs to be
proof,  just simple cite a linear algebra textbook as most
researchers did.
The key issue with which we are concerned is the case of  $\tilde{N}<N$, which means that the number of hidden neurons is smaller than that of training samples.  According the theorem in linear algebra textbook, for example,  Ref. [6] listed in \cite{Guo1995},  as stated in section \ref{leN}, it is impossible we can obtain right inverse of H.  In other words, when we set an infinite small positive value $\epsilon$, the learning error cannot be smaller than this $\epsilon$  when the number of hidden neurons is less than N. 

For a finite N, we know that the pseudoinverse solution is the best approach for output weight matrix W, $W = H^+ T$. If we substitute $W = H^+ T$  into SSE function, we can find that learning problem becomes minimizing $\|PT-T\|$, where $P=HH^+$ is an orthogonal projection operator. If $\tilde{N}<N$ there exists null space for output matrix T, for most training data set, the norm of those vectors which lay in null space for T  cannot be less than $\epsilon$. Furthermore,  in linear algebra the least squares solution for over determined problem has been studied by many researchers, no theory can guarantee  that the error can be arbitrary small for most data set. Therefore, theorem 2.2 is incorrect either.

More theoretical analysis about incorrectness of these theorems, please refer to \cite{Wang2017}.

\subsection{SSE function and generalization}

As we known, the SSE function is the most used function in neural network research. In mathematical expression, the $L_2$ norm is often adopted for the SSE function.  When a finite size training data is given, there are mainly two categories of approaches to avoid underfitting and overfitting, and hence getting good generalization: One is called model selection, and the other is regularization. In \cite{Huang2006}, it stated in the introduction section, ``{\em Different from traditional learning algorithms the proposed learning algorithm not only tends to reach the smallest training error but also the smallest norm of weights.  Therefore, the proposed learning algorithm tends to have good generalization performance for feedforward neural networks}''.  It also claimed in the discussion section of \cite{Huang2006}, ``{\em (2) The proposed ELM has better generalization performance than the gradient-based learning such as back propagation in most cases}''.  We know that when minimizing SSE function (Eq. \ref{sse2}) corresponding to weight W,  pseudoinverse solution is the best approach. This solution has the properties such as the minimum training error,  the smallest norm of weights, and uniqueness of the minimum norm least-squares solution. The ELM authors have misunderstood the meaning of the smallest norm of weights for the pseudoinverse solution. Here the smallest norm of weights is only compared with other least-squares solutions, it cannot guarantee network has good generalization performance if overfitting occurs. To avoid overfitting and obtain good generalization, one of the techniques is weight decay regularization which  adds a penalty term to the SSE function. This constraint term usually is the norm of weights times a regularization constant, as most researchers, including Guo {\em et al} \cite{Guo2003a},  studied in the literature.  In the case SSE function is adopted, other techniques to reach good generalization performance of the neural network include early stop, stacked generalization as discussed in \cite{Guo2004}.  But the studies with different cost functions or regularization techniques to reach good generalization is beyond the scope of this letter. Here we simply wish to point out that if we only minimize the SSE function without any other constraints, it is impossible to get good generalization unless network{'}s architecture is well designed for a given data set. 

\subsection{Hidden Nodes} 
 
In the abstract of Ref.  \cite{Huang2006}, there is a such statement: ``{\em This paper proposes a new learning algorithm called extreme learning machine (ELM) for single-hidden layer feedforward neural networks (SLFNs) which randomly chooses hidden nodes and analytically determines the output weights of SLFNs}''.   We think here the phrase ``randomly chooses hidden nodes'' means randomly chooses the number of hidden nodes,  and with this randomly chosen value, to construct a SLFN.  This selection method for the number of hidden nodes is simple indeed, but it does NOT work at all in practice.  As we know, in order to obtain the good generalization performance for a SLFN for a given data set, one of the techniques is model selection, that is, selecting an optimal network architecture,  a structure neither too simple to underfitting, nor too complex to overfitting.  For a SHLN, it only has one hyperparamenter of the network structure, which is the number of hidden nodes when we assigned the numbers of input and output neurons.  In the past decades, many research papers address this model selection problem, and we get to know that this hyperparameter depends on many factors \cite{NNFAQ}. It is a pitfall  to realize good generalization  through randomly  choosing a number of  hidden nodes.   Let us do a thought experiment to illustrate the random nodes method is useless.  Suppose that a given data set has N training samples, there are N users using this method to do experiments.  Every user chooses randomly a number of hidden nodes in the range of [1, N], which leads to many different network structures being generated. For those users with $\tilde{N}\ll N$, they will suffer from underfitting problem. For those users with $\tilde{N}\sim N$, they may suffer from overfitting problem.  Both underfitting and overfitting will have poor generalization. Only a small number of lucky users can get good generalization performance for their experiments. This thought experiment shows that the number of hidden nodes should be sophisticatedly chosen, instead of merely randomly chosen. Furthermore, in the experiments of \cite{Huang2006},  it can be found that  the cross-validation method was adopted to choose the optimal number of hidden nodes. This confirms that not only their statements have contradicted themselves, but also learning speed of ELM is NOT so fast as they claimed when selecting optimal number of hidden nodes time is counted.

\section{Summary}

In order to stress the originality of the so called ELM,  the ELM authors made such a statement in \cite{Huang2006}, ``{\em It should be noted that the input weights (linking the input layer to the first hidden layer) and hidden layer biases\footnote{The concept is wrong that regards $b_1$ as hidden layer bias, in fact it is input layer bias.} need to be adjusted in all these previous theoretical research works as well as in almost all practical learning algorithms of feedforward neural networks.}''  However,  the fact is that in the PIL,  all three methods to set input weights have shown that weight parameters do not need to be adjusted further.  Also, the number of hidden neurons (including bias neuron) is set to $N$  and does not need to be adjusted either.  Here it is clearly shown that ELM authorÕs statement is a false claim.  Furthermore, During the period of 2005 international conference on intelligent computing (ICICÕ2005), the first author of the PIL papers had introduced PIL work to the first author of the ELM papers.  ELM authors already got known the PIL work, they still wrote such a statement in their paper.  They not only excluded PIL papers in reference list of the article \cite{Huang2006}, but also denied the originality of the PIL. 

     From   discussions in this manuscript, we can see that the  ELM is the same with PIL0 in learning scheme, and ELM not only has nothing new in learning scheme, but is also riddled with a lot of incorrect statements and false claims in theory. To avoid misleading junior researchers around world, our suggestions to the ELM authors are as follows:
\begin{enumerate}
\item Acknowledge the originality of the PIL fast learning scheme and clarify that the ELM is simply a VEST of the PIL algorithm for SHLN.
\item Remedy those incorrect statements and false claims in ELM papers.
\end{enumerate}



\section*{Acknowledgment}


     We greatly appreciate those researchers who provide many useful suggestions and comments to this letter.  Without their help, we still did not know PIL has been renamed to ELM. And we especially thank Prof. Philip C. L. Chen, our
co-author of the PIL, who gives a lot of suggestions, including some key issues, to this work.

%\bibliography{VEST_ref}   % name your BibTeX data base

\appendix

%\section*{Appendix}
About the word ``VEST''.

Firstly,  we would like to express our grateful to Prof. Tom Dietterich, the Moderator for cs.LG at arXiv.  He suggested that ``{\em the word `vest' might not be appropriate. In English, it usually refers either to a type of clothing or to a kind of financial contract. The meaning of `synonym' or `alias' is not known to me,}''  and ``{\em English speakers will be very confused}'' 
by the word.   His kindly suggestion has promoted us to search
a proper word, now we create an abbreviation word ``VEST'', and hope that it  has the similar meaning with word ``vest'' in Chinese slang.

In this manuscript, the ``VEST''  is the abbreviation  of  the \textcolor{blue}{\underline v}ariant cr\textcolor{blue}{\underline e}ated by \textcolor{blue}{\underline s}imple name al\textcolor{blue}{\underline t}ernation. 

Original word ``vest'',  (MaJia in Chinese), is a garment worn on the upper body and must be close fitting  (https://en. wikipedia.org/wiki/Vest). Currently, in Chinese slang it means that one person or thing has various appearances. The origin of this word's slang meaning is from Zhao Benshan and Song DandanÕs opusculum ``the hour worker'' in 2000: ``Childish guy, do not think that I can not recognize you after you wear a vest.''

\end{document}